\documentclass[letterpaper]{article}
\usepackage{aaai}
\usepackage{times}
\usepackage{helvet}
\usepackage{courier}
\usepackage{natbib}
\usepackage{tabularx}
\usepackage{graphicx}
\usepackage{amsmath}
\usepackage{amssymb}
\usepackage{color}
\usepackage{pifont}
\usepackage{bm}
\usepackage{tablefootnote}
\usepackage[textsize=scriptsize]{todonotes}
\definecolor{blue}{RGB}{0, 93, 170}			%Go Big Blue!

\aclfinalcopy
\begin{document}
\title{Exploring Graph-structured Passage Representation for Multi-hop Reading Comprehension with Graph Neural Networks}
\author{Linfeng Song$^1\thanks{~~~Work done during an internship at IBM Research.}$, Zhiguo Wang$^2$\thanks{~~~Co-mentoring.}, Mo Yu$^{2\dagger}$, Yue Zhang$^3$, Radu Florian$^2$ \and Daniel Gildea$^1$ \\
$^1$University of Rochester, Rochester, NY, 14623\\
$^2$IBM T.J. Watson Research Center, Yorktown Heights, NY 10598\\
$^3$School of Engineering, Westlake University, China\\
}

\maketitle
\begin{abstract}
Multi-hop reading comprehension focuses on one type of factoid question, where a system needs to properly integrate multiple pieces of evidence to correctly answer a question. 
Previous work approximates global evidence with local coreference information, encoding coreference chains with DAG-styled GRU layers within a gated-attention reader.
However, coreference is limited in providing information for rich inference.
We introduce a new method for better connecting global evidence, which forms more complex graphs compared to DAGs.
To perform evidence integration on our graphs, we investigate two recent graph neural networks, namely graph convolutional network (GCN) and graph recurrent network (GRN).
Experiments on two standard datasets show that richer global information leads to better answers.
Our method performs better than all published results on these datasets.
\end{abstract}

\section{Introduction}
\noindent Recent years have witnessed a growing interest in the task of machine reading comprehension.
%Several large-scale datasets \citep{rajpurkar-EtAl:2016:EMNLP2016,nguyen2016ms} have been introduced with increasingly promising results \citep{wang2017gated,shen2017reasonet,tan2017s} being reported.
However, most existing work \citep{hermann2015teaching,wang2016machine,seo2016bidirectional,wang2016multi,weissenborn2017making,dhingra-EtAl:2017:Long2,shen2017reasonet} focuses on a factoid scenario where the questions can be answered by simply considering very local information, such as one or two sentences.
For example, to correctly answer a question ``What causes precipitation to fall?'',
a QA system only needs to refer to one sentence in a passage: ``... In meteorology, precipitation is any product of the condensation of atmospheric water vapor that falls under gravity. ...''

A more challenging yet practical extension is the multi-hop reading comprehension (MHRC) \citep{welbl2018constructing}, where a system needs to properly integrate multiple evidence to correctly answer a question.
Figure \ref{fig:example} shows an example, which has three passages, a question and several candidate choices. 
In order to correctly answer the question, a system has to integrate the facts ``The Hanging Gardens are in Mumbai'' and ``Mumbai is a city in India''.
There are also some irrelevant facts, such as ``The Hanging Gardens provide sunset views over the Arabian Sea'' and ``The Arabian Sea is bounded by Pakistan and Iran'', which make the task more challenging, as an MHRC model has to distinguish the relevant facts from the irrelevant ones.

Despite being a practical task, so far MHRC has received relatively little research attention.
One notable method, Coref-GRU \citep{N18-2007}, integrates multiple evidence associated with each entity mention by incorporating coreference information using a collection of GRU layers of a gated-attention reader \citep{dhingra-EtAl:2017:Long2}.
%By modeling coreference structures as directed acyclic graphs using a novel recurrent neural network structure, their method is in line with recent research on lattice-LSTM \citep{su2017lattice,P18-1144} and DAG-LSTM \citep{zhu-sobhani-guo:2016:N16-1,TACL1028}.
However, the main disadvantage of Coref-GRU is that the coreferences it considers are usually local to a sentence, neglecting other useful global information.
The top part of Figure \ref{fig:coref_vs_graph} shows a directed acyclic graph (DAG) with only coreference edges. 
In particular, the two coreference edges indicate the two facts: ``The Hanging Gardens provide views over the Arabian Sea'' and ``Mumbai is a city in India'', from which we cannot indicate the ultimate fact, ``The Hanging Gardens are in India'', for correctly answering this instance.

\begin{figure} 
\begin{tabularx}{0.48\textwidth}{|X|}
\hline
[\textbf{The Hanging Gardens}], in [\textbf{Mumbai}], also known as Pherozeshah Mehta Gardens, are terraced gardens ... [\textbf{They}] provide sunset views over the [\textbf{Arabian Sea}] ... \\
\hline
\hline
[\textbf{Mumbai}] (also known as Bombay, the official name until 1995) is the capital city of the Indian state of Maharashtra. [\textbf{It}] is the most populous city in [\textbf{India}] ... \\
\hline
\hline
The [\textbf{Arabian Sea}] is a region of the northern Indian Ocean bounded 
on the north by [\textbf{Pakistan}] and [\textbf{Iran}], on the west by northeastern [\textbf{Somalia}] and the Arabian Peninsula, and on the east by [\textbf{India}] ... \\
\hline
\textbf{Q}: (The Hanging gardens, country, ?)  \\
\textbf{Options}:  {Iran, India, Pakistan, Somalia, ...} \\
\hline
\end{tabularx}
\vspace{-1.0em}
\caption{An example from WikiHop \citep{welbl2018constructing}, where some relevant entity mentions and their coreferences are highlighted.}
\label{fig:example}
\vspace{-1.6em}
\end{figure}

In this paper, we propose a new approach for evidence integration by considering two more useful types of edges in addition to coreferences.
The bottom part of Figure \ref{fig:coref_vs_graph} shows one example graph generated by our approach.
In particular, we consider three types of edges.
The first type of edges connect the mentions of the \emph{same} entity appearing across passages or further apart in the same passage.
Shown in Figure \ref{fig:coref_vs_graph}, one instance connects the two ``Mumbai'' across the two passages.
Intuitively, \emph{same}-typed edges help to integrate global evidence related to the same entity, which are not covered by pronouns.
The second type of edges connect two mentions of different entities within a context \emph{window}.
They help to pass useful evidence further across entities.
For example, in the bottom graph of Figure \ref{fig:coref_vs_graph}, both \emph{window}-typed edges of \textcircled{1} and \textcircled{6} help to pass evidence from ``The Hanging Gardens'' to ``India'', the answer of this instance.
Besides, \emph{window}-typed edges enhance the relations between local mentions that can be missed by the sequential encoding baseline.
Finally, \emph{coreference}-typed edges are further complimentary to the previous two types.%, and thus we also include them.

With three types of edges, our generated graphs are complex and can have cycles, making it difficult to directly apply a DAG network (e.g. the structure of Coref-GRU).
In addition, certain groups of nodes cannot be reached from each other through graph edges.
Take Figure \ref{fig:coref_vs_graph} as an example. 
Information of ``They'' and ``Arabian Sea'' in the first passage cannot reach ``Mumbai'', ``It'' or ``India'' in the second passage, or vice versa.
To handle these problems, we adopt graph neural networks \citep{scarselli2009graph}, which can encode arbitrary graphs.
In particular, we choose graph convolutional network (GCN) and graph recurrent network (GRN), as they have been shown successful on encoding semantic graphs \citep{P18-1150}, dependency graphs \citep{bastings-EtAl:2017:EMNLP2017,marcheggiani-titov:2017:EMNLP2017,song2018nary} and raw texts \citep{P18-1030}.

\begin{figure}
\centering
\includegraphics[width=0.95\linewidth]{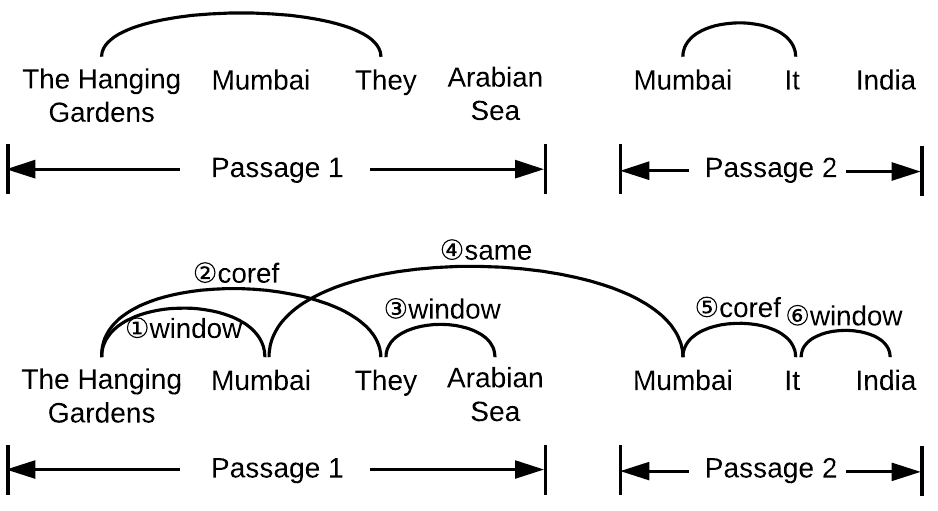}
\vspace{-0.5em}
\caption{A DAG generated by \citet{N18-2007} (top) and a graph by considering all three types of edges (bottom) on the example in Figure \ref{fig:example}.}
\label{fig:coref_vs_graph}
\vspace{-1.0em}
\end{figure}

Given an instance containing several passages and a list of candidates, we first use NER and coreference resolution tools to obtain entity mentions, and then create a graph out of the mentions and relevant pronouns.
As the next step, evidence integration is executed on the graph by adopting a graph neural network on top of a sequential layer.
The sequential layer learns local representation for each mention, while the graph network learns a global representation.
The answer is decided by matching the representations of the mentions against the question representation.

Experiments on WikiHop \citep{welbl2018constructing} and ComplexWebQuestions \citep{N18-1059} show that the additional types of edges we introduced are highly useful for MHRC.
%Our method gives much better performance than our baselines.
On the holdout testset of WikiHop, it achieves an accuracy of 65.4\%, which is the best over all published results on the leaderboard\footnote{http://qangaroo.cs.ucl.ac.uk/leaderboard.html} as of the paper submission time.
On the testset of ComplexWebQuestions, it also achieves better numbers than all published results without additional annotations.
To our knowledge, we are 
among 
the first to investigate graph neural networks on reading comprehension\footnote{The concurrent unpublished work \cite{gcnwiki} also investigate the usage of graph convolution networks on WikiHop. Our work proposes a different model architecture, and focus more on the exploration and comparison of multiple edge types for building the graph-structured passage representation.}.
%We will release our code on http://github.com/xxx after this paper is accepted.

\begin{figure}
\centering
\includegraphics[width=0.95\linewidth]{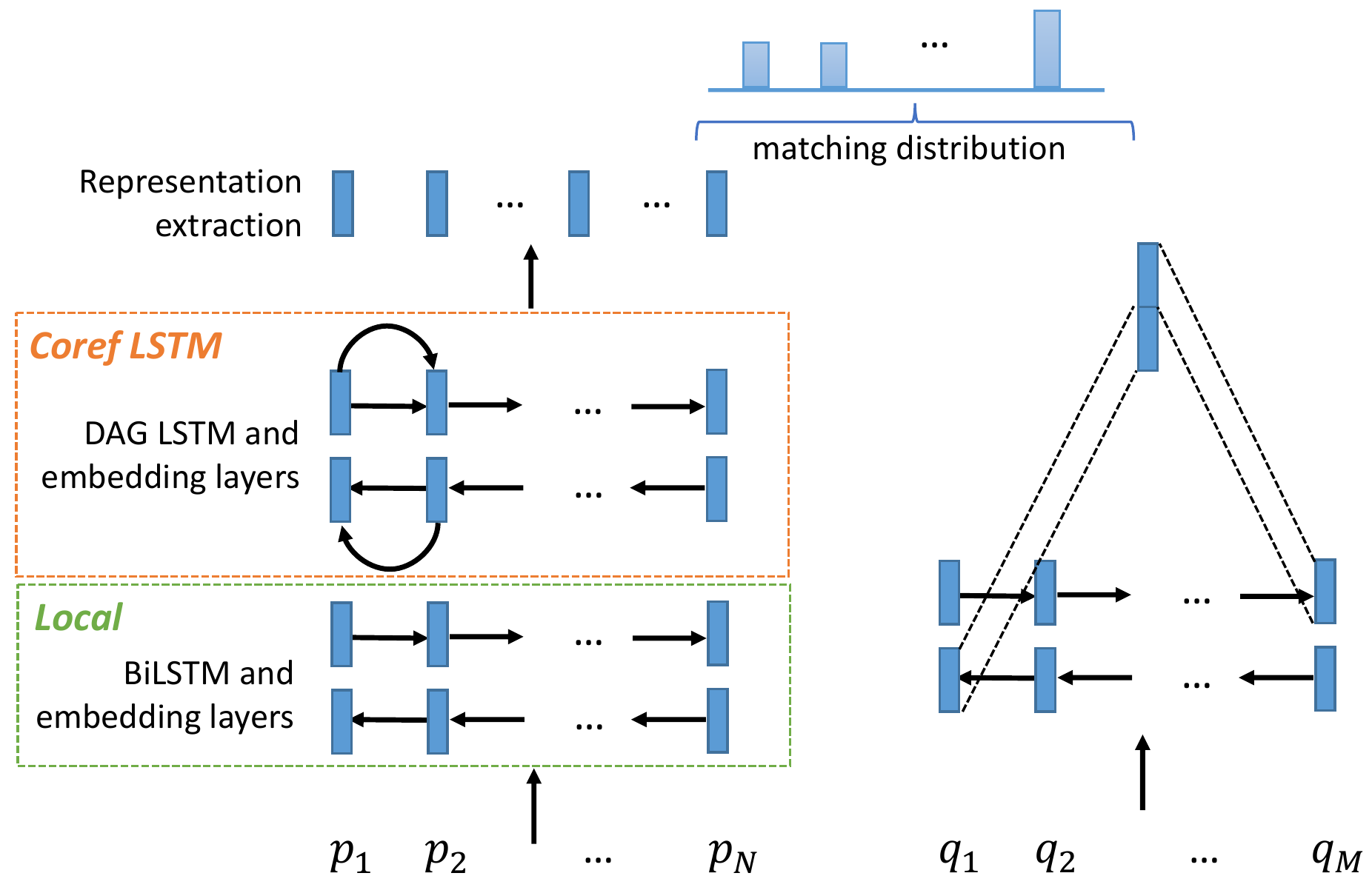}
\vspace{-0.5em}
\caption{Baselines. The upper dotted box is a DAG LSTM layer with addition coreference links, while the bottom is a typical BiLSTM layer. Either layer is used.}
\label{fig:baseline}
\vspace{-1.0em}
\end{figure}

\section{Baseline}

As shown in Figure \ref{fig:baseline}, we introduce two baselines, which are inspired by \citet{N18-2007}.
The first baseline, \emph{Local}, uses a standard BiLSTM layer (shown in the green dotted box), where inputs are first encoded with a BiLSTM layer, then the representation vectors for the mentions in the passages are extracted, before being matched against the question for selecting an answer.
The second baseline, \emph{Coref LSTM}, differs from \emph{Local} by replacing the BiLSTM layer with a DAG LSTM layer (shown in the orange dotted box) for encoding additional coreference information, as proposed by \citet{N18-2007}.

%We build our model on top of the BiLSTM baseline for better studying the behavior of our model.
%The DAG LSTM is for studying coreference within our framework\mo{do we need to keep this paragraph?}.

\subsection{\emph{Local}: BiLSTM encoding}

Given a list of relevant passages, we first concatenate them into one large passage $p_1, p_2 \dots p_N$, where each $p_i$ is a passage word and $\boldsymbol{x}_{p_i}$ is the embedding of it.
It adopts a Bi-LSTM to encode the passage:
\begin{align*}
\overleftarrow{\boldsymbol{h}}_p^i &= \textrm{LSTM}(\overleftarrow{\boldsymbol{h}}_p^{i+1}, \boldsymbol{x}_{p_i}) \\
\overrightarrow{\boldsymbol{h}}_p^i &= \textrm{LSTM}(\overrightarrow{\boldsymbol{h}}_p^{i-1}, \boldsymbol{x}_{p_i})
\end{align*}
Each hidden state contains the information of its local context.
Similarly, the question words $q_1, q_2 \dots q_M$ are first converted into embeddings $\boldsymbol{x}_{q_1}, \boldsymbol{x}_{q_2} \dots \boldsymbol{x}_{q_M}$ before being encoded by another BiLSTM:
\begin{align*}
\overleftarrow{\boldsymbol{h}}_q^j &= \textrm{LSTM}(\overleftarrow{\boldsymbol{h}}_q^{j+1}, \boldsymbol{x}_{q_j}) \\
\overrightarrow{\boldsymbol{h}}_q^j &= \textrm{LSTM}(\overrightarrow{\boldsymbol{h}}_q^{j-1}, \boldsymbol{x}_{q_j})
\end{align*}

\subsection{\emph{Coref LSTM}: DAG LSTM with conference}

Taking the passage word embeddings $\boldsymbol{x}_{p_1}, \dots \boldsymbol{x}_{p_N}$ and coreference information as the input, 
the DAG LSTM layer encodes each input word embedding (such as $\boldsymbol{x}_{p_j}$) with the following gated operations\footnote{Only the forward direction is shown for space consideration.}:
\begin{equation*}
\begin{split}
\boldsymbol{i}_j &= \sigma(\boldsymbol{W}_i \boldsymbol{x}_{p_j} + \boldsymbol{U}_i \sum_{i\in \mathbb{N}(j)}\overrightarrow{\boldsymbol{h}}_p^i + \boldsymbol{b}_i) \\
\boldsymbol{o}_j &= \sigma(\boldsymbol{W}_o \boldsymbol{x}_{p_j} + \boldsymbol{U}_o \sum_{i\in \mathbb{N}(j)}\overrightarrow{\boldsymbol{h}}_p^i + \boldsymbol{b}_o) \\
\boldsymbol{f}_{i,j} &= \sigma(\boldsymbol{W}_f \boldsymbol{x}_{p_j} + \boldsymbol{U}_f \overrightarrow{\boldsymbol{h}}_p^i + \boldsymbol{b}_f) \\
\boldsymbol{u}_j &= \sigma(\boldsymbol{W}_u \boldsymbol{x}_{p_j} + \boldsymbol{U}_u \sum_{i\in \mathbb{N}(j)}\overrightarrow{\boldsymbol{h}}_p^i + \boldsymbol{b}_u) \\
\overrightarrow{\boldsymbol{c}}_p^j &= \boldsymbol{i}_j \odot \boldsymbol{u}_j + \sum_{i\in \mathbb{N}(j)} \boldsymbol{f}_{i,j} \odot \overrightarrow{\boldsymbol{c}}_p^i \\
\overrightarrow{\boldsymbol{h}}_p^j &= \boldsymbol{o}_j \odot \tanh (\overrightarrow{\boldsymbol{c}}_p^j) \\
\end{split}
\end{equation*}
$\mathbb{N}_j$ represents all preceding words of $p_j$ in the DAG, $\boldsymbol{i}_j$, $\boldsymbol{o}_j$ and $\boldsymbol{f}_{i,j}$ are the input, output and forget gates, respectively. $\boldsymbol{W}_x$, $\boldsymbol{U}_x$ and $\boldsymbol{b}_x$ ($x \in \{i,o,f,u\}$) are model parameters.

\subsection{Representation extraction}

After encoding both the passage and the question, we obtain a representation vector for each entity mention $\epsilon_k$, spanning from $k_i$ to $k_j$, by concatenating the hidden states of its start and end positions, before they are correlated with a fully connected layer:
\begin{equation} \label{eq:base_rep}
\boldsymbol{h}_\epsilon^k = \boldsymbol{W}_1 [\overleftarrow{\boldsymbol{h}}_p^{k_i};\overrightarrow{\boldsymbol{h}}_p^{k_i};\overleftarrow{\boldsymbol{h}}_p^{k_j};\overrightarrow{\boldsymbol{h}}_p^{k_j}]+\boldsymbol{b}_1 \textrm{,}
\end{equation}
where $W_1$ and $b_1$ are model parameters for compressing the concatenated vector.
Note that the current multi-hop reading comprehension datasets all focus on the situation where the answer is a named entity mention.
Similarly, the representation vector for the question is generated by concatenating the hidden states of its first and last positions:
\begin{equation}
\boldsymbol{h}_q = \boldsymbol{W}_2 [\overleftarrow{\boldsymbol{h}}_q^1;\overrightarrow{\boldsymbol{h}}_q^1;\overleftarrow{\boldsymbol{h}}_q^M;\overrightarrow{\boldsymbol{h}}_q^M]+\boldsymbol{b}_2
\end{equation}
where $W_2$ and $b_2$ are also model parameters.

\subsection{Attention-based matching}

After obtaining the representation vectors for the question and the entity mentions in the passages, an additive attention model \citep{bahdanau2015neural}
%\footnote{We adopt a standard matching method, as our focus is to show the effectiveness of evidence integration. We leave investigating other matching approaches \citep{luong-pham-manning:2015:EMNLP,Wang:2017:BMM:3171837.3171865} as a future work.} 
is adopted by treating all entity mention representations and the question representation as the memory and the query, respectively.
In particular, the probability for an entity $\epsilon$ being the answer is calculated by summing up all the occurrences of $\epsilon$ across the input passages:
\begin{equation} \label{eq:merge}
Pr_\epsilon = \frac{\sum_{k\in \mathcal{N}_\epsilon} \alpha_k}{\sum_{k'\in \mathcal{N}} \alpha_{k'}} \textrm{,}
\end{equation}
where $\mathcal{N}_\epsilon$ and $\mathcal{N}$ represent all occurrences of entity $\epsilon$ and all occurrences of all entities, respectively. 
Previous work \citep{wang2018r3} shows that summing the probabilities over all occurrences of the same entity mention is important for the multi-passage scenario.
$\alpha_k$ is the attention score for the entity mention $\epsilon_k$, calculated by an additive attention model shown below:
\begin{align}
e_0^k &= \boldsymbol{v}_a^T \tanh(\boldsymbol{W}_a \boldsymbol{h}_\epsilon^k + \boldsymbol{U}_a \boldsymbol{h}_q + \boldsymbol{b}_a) \\
a_k &= \frac{\exp(e_0^k)}{\sum_{k'\in \mathcal{N}} \exp(e_0^{k'})} \label{eq:softmax}
\end{align}
where $\boldsymbol{v}_a$, $\boldsymbol{W}_a$, $\boldsymbol{U}_a$ and $\boldsymbol{b}_a$ are model parameters.

\paragraph{Comparison with \citet{N18-2007}}
% \citet{N18-2007} uses a gated-attention reader (GA reader) \citep{dhingra-EtAl:2017:Long2} as their baseline.
% It is designed for the cloze-style reading comprehension task \citep{hermann2015teaching} where \emph{one} token is selected from the input passages as the answer for each instance.
% The GA reader takes several standard GRU layers, and the main contribution of \citet{N18-2007} is replacing these GRU layers with Coref-GRU layers, a kind of DAG network, for capturing coreference information.
% \mo{move the above commented paragraph to the beginning of the section.}
The Coref-GRU model \citep{N18-2007} is based on the gated-attention reader (GA reader) \citep{dhingra-EtAl:2017:Long2}, which is designed for the cloze-style reading comprehension task \citep{hermann2015teaching}, where \emph{one} token is selected from the input passages as the answer for each instance.
To adapt their model for the WikiHop benchmark, where an answer candidate can contain multiple tokens, they first generate a probability distribution over the passage tokens with GA reader, and then compute the probability for each candidate $c$ by aggregating the probabilities of all passage tokens that appear in $c$ and renormalizing over the candidates.

% After encoding both passages and question, a GA reader calculates a probability distribution over the passage tokens by an attention mechanism \textsc{Attn}:
% \[
% Pr_1 \dots Pr_N = \textrm{softmax}((h_q)^T H_p)
% \]
% where $H_p=[h_p^1\dots h_p^N]$ is the concatenation of the hidden states of all passage words.
% The probability for a candidate $c$ is computed by aggregating the probabilities of all passage tokens that appear in $c$ and renormalizing over the candidates, and then the candidate with the highest probability is picked as the predicted answer:
% \begin{align}
% \textrm{argmax}_{c\in C} \sum_{i\in \Psi_c} p_i
% \end{align}
% where $\Psi_c$ is the set of positions where a token in $c$ appears in the passages.

In addition to using LSTM instead of GRU\footnote{Model architectures are selected according to dev results.}, the main difference between our two baselines and \citet{N18-2007} is that our baselines consider each candidate as a whole unit no matter whether it contains multiple tokens or not. 
This makes our models more effective on the datasets containing phrasal answer candidates.

\begin{figure}
\centering
\includegraphics[width=0.95\linewidth]{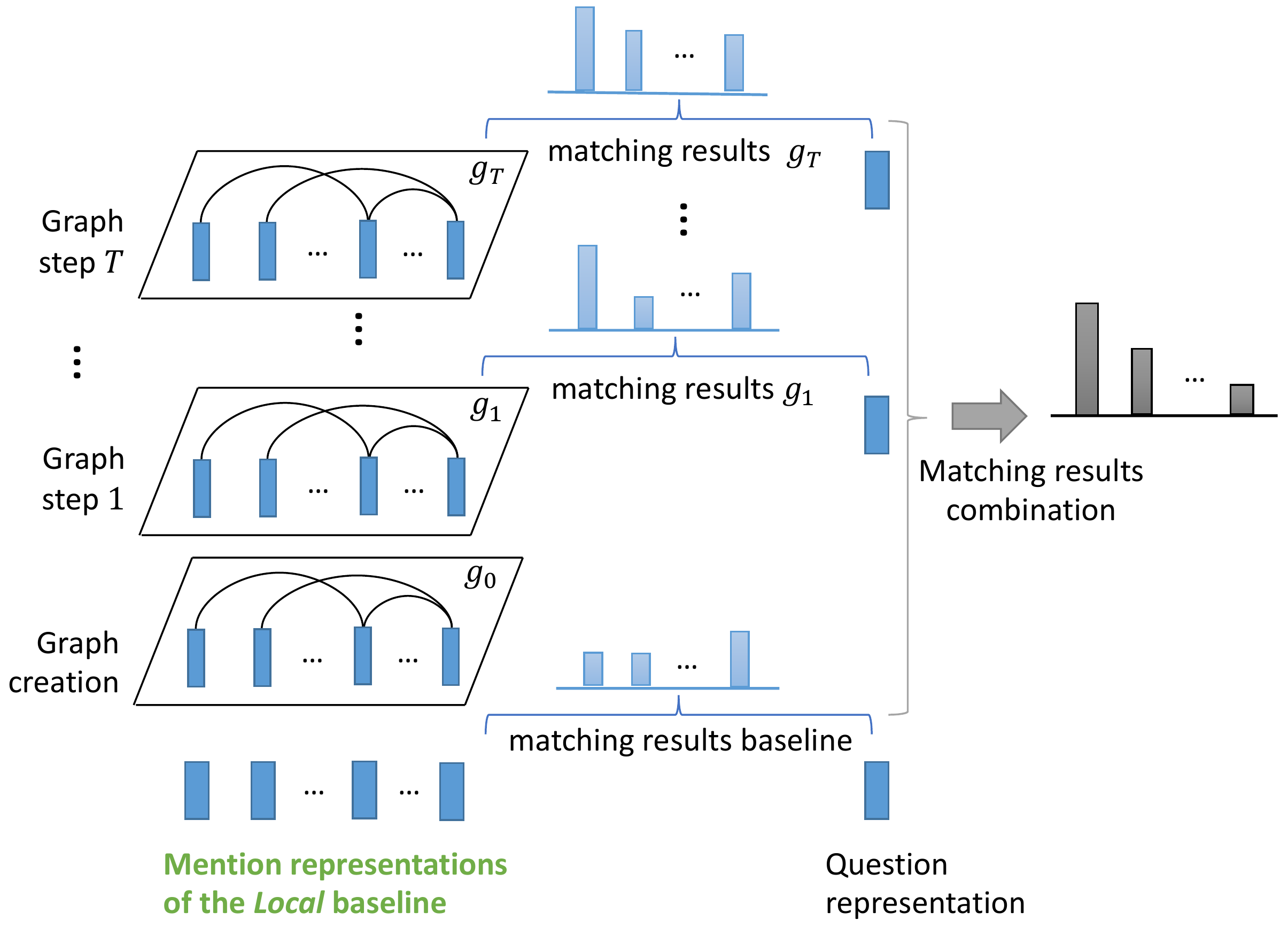}
\vspace{-0.5em}
\caption{Model framework.}
\label{fig:model_overview}
\vspace{-1.0em}
\end{figure}

\section{Evidence integration with graph network}

After obtaining the representation vectors for a question and the corresponding entity mentions, we build a graph out of the entity mentions by connecting relevant mentions with edges, and then integrating relevant information for each graph node (entity mention) with a graph recurrent network (GRN) \citep{P18-1030,P18-1150} or a graph convolutional network (GCN) \citep{kipf2017semi}.
Figure \ref{fig:model_overview} shows the overall procedure.

\subsection{Graph construction}

As a first step, we create a graph from a list of input passages.
The entity mentions within the passages are taken as the graph nodes. 
They are automatically generated by NER and coreference annotators, so that each graph node is either an entity mention or a pronoun representing an entity.
We then create a graph by ensuring that edges between two nodes follow the situations below:
\begin{itemize}
\item They are occurrences of the \textbf{same} entity mention across passages or with a distance larger than a threshold $\tau_L$ when being in the same passage.
\item One is an entity mention and the other is its \textbf{coreference}. Our coreference information is automatically generated by a coreference annotator.
\item Between two mentions of different entities in the same passage within a \textbf{window} threshold of $\tau_S$.
\end{itemize}
Between every two entities that satisfy the situations above, we make two edges in opposite directions. 
As a result, each generated graph can also be considered as an undirected graph.

%\mo{Can be commented if space isn't enough}
%As mentioned in the introduction, each type of edges capture a specific correlation.
%The \emph{same}-typed edges link all the mentions of an entity, and the \emph{coreference}-typed edges connect each mention with its coreferences.
%Both type of edges help integrate evidence of the same entity.
%On the other hand, \emph{window}-typed edges help to merge evidence of different entities and to pass evidence further in the graph.
%For example in Figure \ref{fig:coref_vs_graph}, two \emph{window}-typed edges help to pass evidence related to ``The Hanging Gardens'' to ``India'', which is the final answer.
%In addition to passing evidence further, \emph{window}-typed edges can enhance the relations between local mentions.
%Such information may not be properly encoded (or even dropped) by bidirectional LSTM encoding, thus additional relational information is useful in this case \citep{marcheggiani2018exploiting}.
%We perform an ablation study of the three types of edges in our experiments.

\subsection{Evidence integration with graph network}

%After creating a graph, the initial state of a graph node only contains the local information of an entity mention.
%As the next step, we investigate two recent graph networks, GRN and GCN, to incorporate other relevant information.

Tackling multi-hop reading comprehension requires inferring on global context.
As the next step, we merge related information through the three types of edges just created. 
We investigate two recent graph networks: GRN and GCN.

\vspace{0.5em}
\textbf{Graph recurrent network (GRN)} 
GRN
%\citep{P18-1030,P18-1150} 
models a graph as a single state, performing recurrent information exchange between graph nodes through graph state transitions.
Formally, given a graph $G=(V,E)$, a hidden state vector $s^k$ is created to represent each node $v_k \in V$.
The state of the graph can thus be represented as:
\[
\boldsymbol{g}=\{\boldsymbol{s}^k\}|v_k \in V
\]
In order to integrate non-local evidence among nodes, information exchange between neighborhooding nodes is performed through recurrent state transitions, leading to a sequence of graph states $\boldsymbol{g}_0, \boldsymbol{g}_1, \dots, \boldsymbol{g}_t$, where $\boldsymbol{g}_t=\{\boldsymbol{s}^k_t\}|v_k \in V$ and $t$ is a hyperparameter representing the number of graph state transition decided by a development experiment.
For initial state $\boldsymbol{g}_0=\{\boldsymbol{s}^k_0\}|v_k \in V$, we initialize each $\boldsymbol{s}^k_0$ by:
\begin{equation} \label{eq:init}
\boldsymbol{s}^k_0 = \boldsymbol{W}_3 [\boldsymbol{h}_\epsilon^k; \boldsymbol{h}_q] + \boldsymbol{b}_3 \textrm{,}
\end{equation}
where $\boldsymbol{h}_\epsilon^k$ is the corresponding representation vector of entity mention $v_k$, calculated by Equation \ref{eq:base_rep}. $\boldsymbol{h}_q$ is the question representation. $\boldsymbol{W}_3$ and $\boldsymbol{b}_3$ are model parameters.

A gated recurrent neural network is used to model the state transition process. 
In particular, the transition from $\boldsymbol{g}_{t-1}$ to $\boldsymbol{g}_t$ consists of a hidden state transition for each node.
At each step $t$, direct information exchange is conducted between a node and all its neighbors via the following LSTM \citep{hochreiter1997long} operations:
\begin{equation} \label{eq:grn_mp}
\begin{split}
\boldsymbol{i}_t^k &= \sigma (\boldsymbol{W}_i \boldsymbol{m}_{t}^k + \boldsymbol{b}_i) \\
\boldsymbol{o}_t^k &= \sigma (\boldsymbol{W}_o \boldsymbol{m}_{t}^k + \boldsymbol{b}_o) \\
\boldsymbol{f}_t^k &= \sigma (\boldsymbol{W}_f \boldsymbol{m}_{t}^k + \boldsymbol{b}_f) \\
\boldsymbol{u}_t^k &= \sigma (\boldsymbol{W}_u \boldsymbol{m}_{t}^k + \boldsymbol{b}_u) \\
\boldsymbol{c}_t^k &= \boldsymbol{f}_t^k \odot \boldsymbol{c}_{t-1}^k + \boldsymbol{i}_t^k \odot \boldsymbol{u}_t^k \\
\boldsymbol{s}_t^k &= \boldsymbol{o}_t^k \odot \tanh(\boldsymbol{c}_t^k) \textrm{,}
\end{split}
\end{equation}
where $\boldsymbol{c}_t^k$ is the cell vector to record memory for $\boldsymbol{s}_t^k$, and $\boldsymbol{i}_t^k$, $\boldsymbol{o}_t^k$ and $\boldsymbol{f}_t^k$ are the input, output and forget gates, respectively. 
$\boldsymbol{W}_x$ and $\boldsymbol{b}_x$ ($x\in \{i,o,f,u\}$) are parameters.
$\boldsymbol{m}_{t}^k$ is the sum of the neighborhood hidden states for the node $v_k$\footnote{We tried distinguishing the neighbors connected by different types of edges, but it does not improve the performance.}:
\begin{equation} \label{eq:aggre}
\boldsymbol{m}_{t}^k = \sum_{i \in \mathbb{N}(k)} \boldsymbol{s}_{t-1}^i
\end{equation}
$\mathbb{N}(k)$ represents the set of all neighbors of $v_k$.

\vspace{0.5em}
\textbf{Graph convolutional network (GCN)} 
GCN 
%\citep{kipf2017semi} 
is a convolution-based alternative to GRN for encoding graphs.
%, and its variants have been applied on NLP problems, such as semantic role labeling \citep{marcheggiani-titov:2017:EMNLP2017} and machine translation \citep{bastings-EtAl:2017:EMNLP2017}.
Similar with GRN, a GCN model consists of two main steps: state initialization and state update.
For state initialization, GCN adopts the same approach as with GRN by initializing from the representations vectors of entity mentions, as shown in Equation \ref{eq:init}.
The main difference between GCN and GRN is the way for updating node states.
GRN adopts gated operations (shown in Equation \ref{eq:grn_mp}), while GCN uses linear transportation with sigmoid:
\begin{equation}
\boldsymbol{s}_t^k = \sigma (\boldsymbol{W}_g \boldsymbol{m}_{t}^k + \boldsymbol{b}_g) \textrm{,} 
\end{equation}
where $\boldsymbol{m}_{t}^k$ is also the sum of the neighborhood hidden states defined in Equation \ref{eq:aggre}. 
$\boldsymbol{W}_g$ and $\boldsymbol{b}_g$ are model parameters.

%\mo{Can be commented}Both GCN and GRN can be cast into one unified process of iteratively computing massages from neighbors to the current node, and then updating the node state by applying messages.
%Both GCN and GRN compute messages by summing up the hidden states of neighbors.
%GRN adopts gated operations for updating states with computed messages, while GCN uses convolutional transformation.

\subsection{Matching and combination}

After evidence integration, we match the hidden states at each graph encoding step with the question representation using the same additive attention mechanism introduced in the Baseline section.
In particular, for each entity $v_k$, the matching results for the baseline and each GRN step $t$ are first generated, before being combined using a weighted sum to obtain the overall matching result:
\begin{align}
e_t^k &= \boldsymbol{v}_{a_t}^T \tanh(\boldsymbol{s}_t^k \boldsymbol{W}_{a_t} + \boldsymbol{h}_q \boldsymbol{U}_{a_t} + \boldsymbol{b}_{a_t}) \\
e^k &= \boldsymbol{w}_c \odot [e_0^k, e_1^k, \dots, e_T^k] + b_c \textrm{,}
\end{align}
where $e_0^k$ is the baseline matching result for $v_k$, $e_t^k$ is the matching results after $t$ GRN steps and $T$ is the number of graph encoding steps. 
$\boldsymbol{W}_{a_t}$, $\boldsymbol{U}_{a_t}$, $\boldsymbol{v}_{a_t}$, $\boldsymbol{b}_{a_t}$, $\boldsymbol{w}_c$ and $b_c$ are model parameters.
In addition, a probability distribution is calculated from the overall matching results using softmax, similar to Equations \ref{eq:softmax}. 
Finally, probabilities that belong to the same entity mention are merged to obtain the final distribution, as shown in Equation \ref{eq:merge}.

\section{Training} 
We train both the baseline and our models using the cross-entropy loss:
\[
l = -\log p(\epsilon^*|X;\theta) \textrm{,}
\]
where $\epsilon^*$ is ground-truth answer, $X$ and $\theta$ are the input and model parameters, respectively.
Adam \citep{kingma2014adam} with a learning rate of 0.001 is used as the optimizer. 
Dropout with rate 0.1 and a $l$2 normalization weight of $10^{-8}$ are used during training.

\section{Experiments on WikiHop}

We study the effectiveness of the three types of edges and the graph encoders using WikiHop \citep{welbl2018constructing} dataset.
%It is designed for multi-evidence reasoning, as the construction process makes sure that multiple evidence are required for inducing the answer for each instance.

\subsection{Data}

The dataset contains around 51K instances, including 44K for training, 5K for development and 2.5K for held-out testing.
Each instance consists of a question, a list of associated passages, a list of candidate answers and a correct answer.
One example is shown in Figure \ref{fig:example}.
%On average each instance has around 19 candidates, all of which are the same type.
%For example, if the answer is a country, all other candidates are also countries. 
We use Stanford CoreNLP \citep{manning-EtAl:2014:P14-5} to obtain coreference and NER annotations. 
Then the entity mentions, pronoun coreferences and the provided candidates are taken as graph nodes to create an evidence graph.
The distance thresholds ($\tau_L$ and $\tau_S$, in \textbf{Graph construction}) for making \emph{same} and \emph{window} typed edges are set to 200 and 20, respectively.
%To make edges, we connect two mentions if one of the following conditions are satisfied:
%\begin{itemize}
%\item The mentions correspond to the same entity and are either in different passages or are separated by 200 tokens or more in one passage.
%\item One mention is a pronoun referring to the other.
%\item The mentions correspond to different entities and are in the same passage with a distance of at most 20 tokens.
%\end{itemize}

\begin{figure}
\centering
\includegraphics[width=0.95\linewidth]{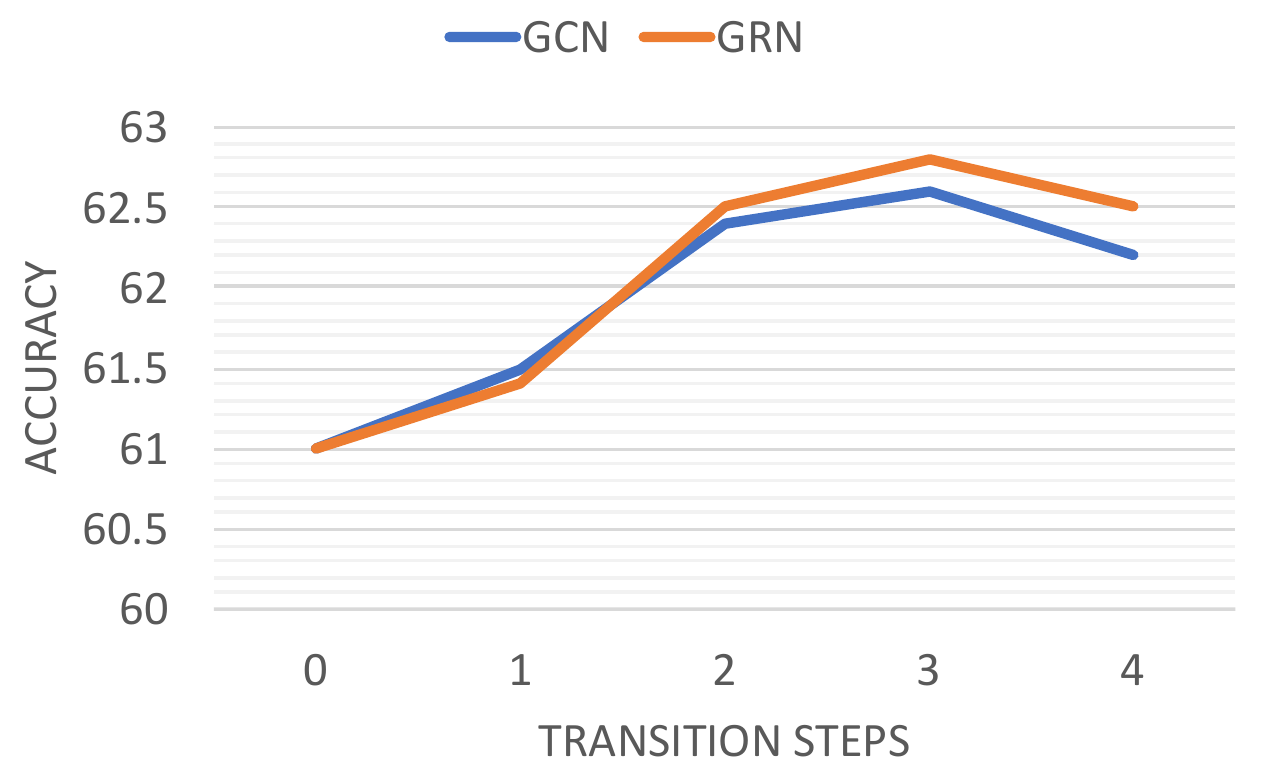}
\caption{\textsc{Dev} performances of different transition steps.}
\label{fig:dev_steps}
\end{figure}

\subsection{Settings}
We study the model behavior on the WikiHop devset, choosing the best hyperparameters for online system evaluation on the final holdout testset.
Our word embeddings are initialized from the 300-dimensional pretrained Glove word embeddings \citep{pennington2014glove} on Common Crawl, and are not updated during training.

For model hyperparameters, we set the graph state transition number as 3 according to development experiments.
Each node takes information from at most 200 neighbors, where \emph{same} and \emph{coref} typed neighbors are kept first. 
The hidden vector sizes for both bidirectional LSTM and GRN layers are set to 300\footnote{We tried larger hidden sizes for our baselines, but did not observe further improvement.}.

\subsection{Development experiments}

%\subparagraph{Transition steps}
Figure \ref{fig:dev_steps} shows the devset performances of our model using GRN or GCN with different transition steps.
It shows the baseline performances when transition step is 0.
The performances go up for both models when increasing the transition step to 3.
Further increasing the transition step leads to a slight decrease in performance.
One reason can be that executing more transition steps may also introduce more noise through richly connected edges.
GRN shows better performances than GCN with large transition steps, indicating that GRN are better at capturing long-range dependency.
This is likely because the gated operations of GRN is better at handling the vanishing/exploding gradient problem than the linear operations of GCN.

\begin{table}
\centering
\begin{tabular}{l|c|c}
Model & Dev & Test \\
\hline
GA w/ GRU & 54.9 & -- \\
GA w/ Coref-GRU & 56.0 & 59.3 \\
\hline
Local & 61.0 & -- \\
Coref LSTM & 61.4 & -- \\
Coref GRN & 61.4 & -- \\
MHQA-GCN & 62.6 & -- \\
MHQA-GRN & \bf 62.8 & \bf 65.4 \\
\end{tabular}
\caption{Main results (unmasked) on WikiHop.}
\label{tab:wikihop}
\end{table}

\subsection{Main results}

Table \ref{tab:wikihop} shows the main comparison results with existing work, where
\emph{GA w/ GRU} and \emph{GA w/ Coref-GRU} correspond to \citet{N18-2007}, and their reported numbers are copied.
The former is their baseline, a gated-attention reader \citep{dhingra-EtAl:2017:Long2}, and the latter is their proposed method.
\emph{Local} is our baseline encoding input passages with a BiLSTM, which only captures local information for each mention.
\emph{Coref LSTM} is our baseline that encodes input passages with coreference annotations by using a bidirectional DAG LSTM.
This can be considered as a reimplementation of \citet{N18-2007} based on our framework.
\emph{Coref GRN} is another baseline that uses GRN for encoding coreference.
It is an ablation study of our model on coreference DAGs, and is for contrasting a DAG network with a graph network.
\emph{MHQA-GCN} and \emph{MHQA-GRN} correspond to our evidence integration approaches via graph encoding, adopting GCN and GRN for graph encoding, respectively.

\begin{table}
\centering
\begin{tabular}{l|c}
% \toprule
Edge type & Dev \\
\hline
all types & 62.8 \\
~~~~~w/o same & 61.9 \\
~~~~~w/o coref & 61.7 \\
~~~~~w/o window & 62.4 \\
\hline
only same &  61.6 \\
only coref & 61.4 \\
only window & 61.1 \\
% \hline
\end{tabular}
%\vspace{-0.5em}
\caption{Ablation study on different types of edges using GRN as the graph encoder.}
\label{tab:dev_edges}
%\vspace{-1.0em}
\end{table}

Our baselines and models show much higher accuracies compared with \emph{GA w/ GRU} and \emph{GA w/ Coref-GRU}, as our models are more compatible with the evaluated dataset.
In particular, we consider each candidate answer as a single unit, while \emph{GA w/ GRU} and \emph{GA w/ Coref-GRU} calculate the probability for each candidate by summing up the probabilities of all tokens within the candidate. 
%In particular, the gated-attention reader, which their work is based on, is designed for the cloze-style reading comprehension tasks where only one passage token is selected as the answer.
%On the other hand, many candidates of the WikiHop dataset have multiple tokens.
%In order to apply their models, the occurrences of the candidates need to be masked with randomly generated tokens\footnote{WikiHop provides a masked version}, thus important information is lost.

\emph{Coref LSTM} only shows 0.4 points gains over \emph{Local}.
On the other hand, \emph{MHQA-GCN} and \emph{MHQA-GRN} are 1.4 and 1.8 points more accurate than \emph{Local}, respectively.
This is mainly because our graphs are better connected than coreference DAGs and are more suitable for integrating relevant evidence.
%Encoding with GRN gives an increase of 0.2 points compared with GCN, because GRN uses more powerful gated operations that can better remember complex combinations of evidence.
\emph{Coref GRN} gives a comparable performance with \emph{Coref LSTM}, showing that graph networks may not necessarily be better than DAG networks on encoding DAGs.
However, the former are more general on encoding arbitrary graphs.
Finally, \emph{MHQA-GRN} shows a higher testing accuracy than all published results\footnote{At submission time we observe a recent short arXiv paper \citep{gcnwiki}, available on August 28th, showing an accuracy of 67.6 using ELMo \citep{peters2018deep}, which is the only result better than ours.
ELMo has achieved dramatic performance gains of 3+ points over a broad range of tasks.
Our main contribution is studying an evidence integration approach, which is orthogonal to the contribution of ELMo on large-scale training.
For more fair comparison with existing work, we did not adopt ELMo, but we will conduct experiments with ELMo as well.}.

\subsection{Analysis}

\subparagraph{Effectiveness of edge types}
Table \ref{tab:dev_edges} shows the effectiveness of different types of edges that we introduce.
The first group shows the ablation study, which indicates the importance of each type of edges.
Among all these types, removing \emph{window}-typed edges causes the least performance drop.
One possible reason is that some information captured by \emph{window}-typed edges has been well captured by sequential encoding.
However, \emph{window}-typed are still useful, as they can help passing evidence through to further nodes.
Take Figure \ref{fig:coref_vs_graph} as an example, \emph{window}-typed edges help to pass information from ``The Hanging Gardens'' to ``India''.
The other two types of edges are more important than \emph{window}-typed ones.
Intuitively, they help to learn a better representation for an entity by integrating the contextual information from its co-references and occurrences.

The second group of Table \ref{tab:dev_edges} shows the model performances when only one type of edges are used.
The numbers generally demonstrate the same patterns as the first group.
In addition, \emph{only same} is slightly better than \emph{only coref}.
It is likely because some coreference information can also be captured by sequential encoding.
%In particular, it shows that \emph{same} and \emph{coref}-typed edges are more effective than \emph{window}-typed ones.
None of the results with a single edge type is significantly better than our strong baseline, whereas the combination of all three types achieves a much better result. 
This indicates the importance of evidence integration over multiple edge types.

\begin{figure}
\centering
\includegraphics[width=0.85\linewidth]{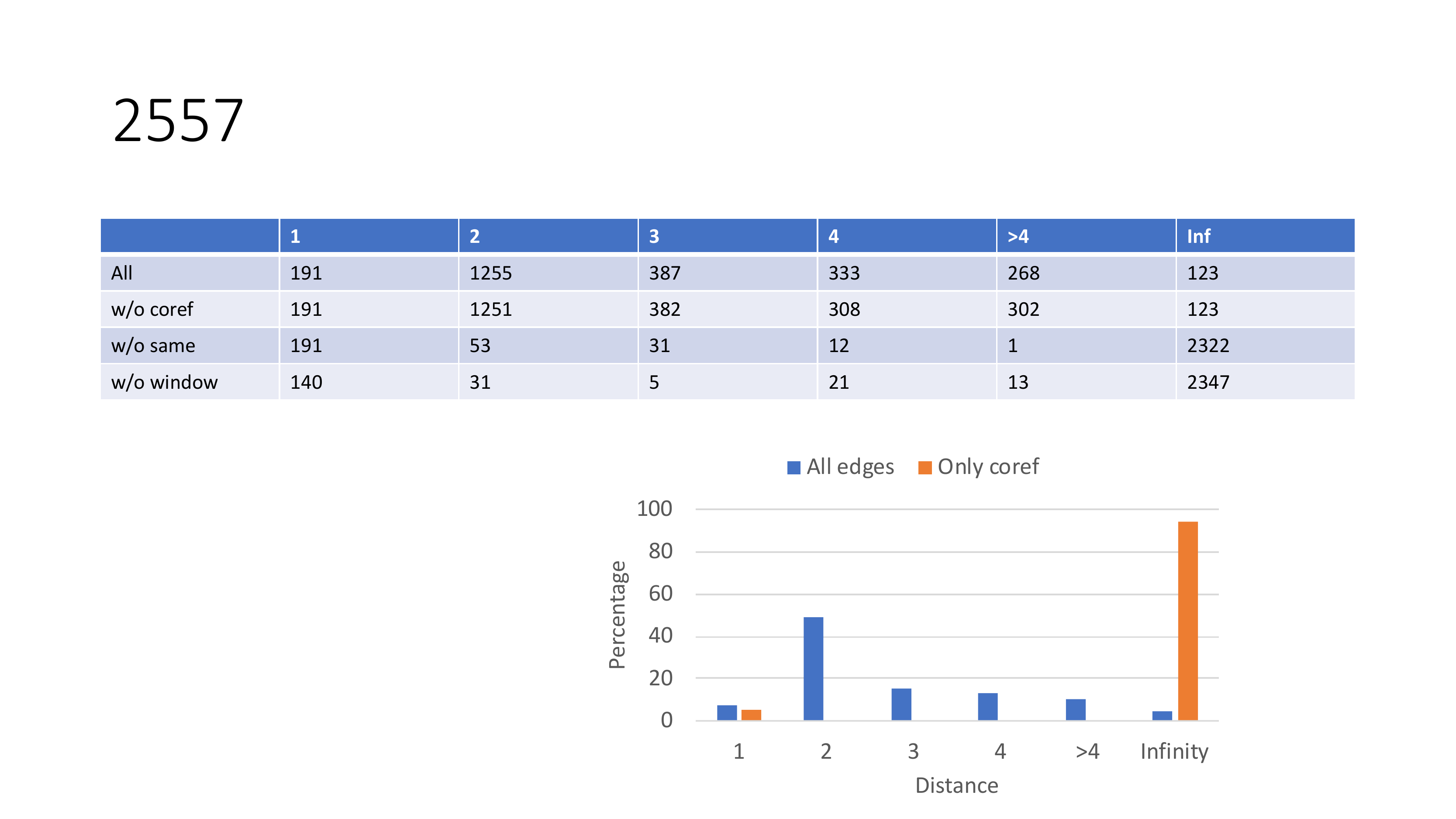}
%\vspace{-1.0em}
\caption{Distribution of distances between a question and an answer on the \textsc{Devset}.}
\label{fig:dist}
%\vspace{-1.0em}
\end{figure}

\vspace{0.5em}
\textbf{Distance}
Figure \ref{fig:dist} shows the percentage distribution of distances between a question and its closet answer when either all types of edges are adopted or only coreference edges are used.
The subject of each question\footnote{As shown in Figure \ref{fig:example}, each question is a three-element tuple of subject, relation and a question mark (asking for the object).} is used to locate the question on the corresponding graph.

When all types of edges are used, the instances with distances less than or equal to 3 count for around 70\% of all the instances. 
On the other hand, the instances with distances longer than 4 only count for 10\%.
This can be the reason why performances do not increase when more than 3 transition steps are performed in our model.
The advantage of our graph construction method can be shown by contrasting the distance distributions over graphs generated by both the baseline and our method.
We further evaluate both methods on a subset of the devset instances, where for each instance the distance between the answer and the question is at most 3 in our graph but is infinity on the coreference DAG.
The performances of \emph{Coref LSTM} and \emph{MHQA-GRN} on this subset are 61.1 and 63.8, respectively.
Comparing with the performances on the whole devset (61.4 vs 62.8), the performance gap is increased by 1.3 points on this subset, which further confirms our observation.
%Even though the question-answer distance becomes infinity for most instances with only coreference edges, please note that the coreference still help to incorporate more context to the entity mentions.
%As a result, coreference edges are also important.

\section{Experiments on ComplexWebQuestions}

In additional to WikiHop, we conduct experiments on the newly released ComplexWebQuestions version 1.1 \citep{N18-1059} for better evaluating our approach.
Compared with WikiHop, where the complexity is implicitly specified in the passages, the complexity of this dataset is explicitly specified on the question side.
One example question is ``What city is the birthplace of the author of `Without end'''.
A two-step reasoning is involved, with the first step being ``the author of `Without end''' and the second being ``the birthplace of $x$''. 
$x$ is the answer of the first step.

\begin{table}
\centering
\begin{tabular}{l|c|c}
Model & Dev & Test \\
\hline
%SimpQA (pretrained RC) & 20.5 & 19.9 \\
%SplitQA (pretrained RC) & 27.6 & 25.9 \\
%\hline
SimpQA & 30.6 & -- \\
SplitQA & 31.1 & -- \\
\hline
Local & 31.2 & 28.1 \\
% \quad+Reranker \citep{wang2018evidence} & 31.0 & -- \\
MHQA-GCN & 32.8 & -- \\
MHQA-GRN & \bf 33.2 & \bf 30.1 \\
MHQA-GRN w/ only same & 32.2 & -- \\
\hline
\hline
SplitQA w/ additional labeled data & 35.6 & 34.2 \\
\end{tabular}
%\vspace{-0.5em}
\caption{Results on the ComplexWebQuestions dataset.}
\label{tab:compqa}
%\vspace{-1.0em}
\end{table}

In this dataset, web snippets (instead of passages as in WikiHop) are used for extracting answers.
The baseline of \citet{N18-1059} (\emph{SimpQA}) only uses a full question to query the web for obtaining relevant snippets, while their model (\emph{SplitQA}) obtains snippets for both the full question and its sub-questions.
With all the snippets, \emph{SplitQA} models the QA process based on a computation tree\footnote{A computation tree is a special type of semantic parse, which has two levels. The first level contains sub-questions and the second level is a composition operation.} of the full question. 
In particular, they first obtain the answers for the sub-questions, and then integrate those answers based on the computation tree.
In contrast, our approach creates a graph from all the snippets, thus the succeeding evidence integration process can join all associated evidence.

\vspace{0.5em}
\textbf{Main results}
As shown in Table \ref{tab:compqa}, similar to the observations in WikiHop, both \emph{MHQA-GRN} and \emph{MHQA-GCN} achieve large improvements over \emph{Local}, and \emph{MHQA-GRN} gives slightly better accuracy.
Both the baselines and our models use all web snippets, but \emph{MHQA-GRN} and \emph{MHQA-GCN} further consider the structural relations among entity mentions.
\emph{SplitQA} achieves 0.5\% improvement over \emph{SimpQA}\footnote{Upon the submission time, the authors of ComplexWebQuestions have not reported testing results for the two methods. To make a fair comparison we compare the devset accuracy.}. 
Our \emph{Local} baseline is comparable with \emph{SplitQA} and our graph-based models contribute a further 2\% improvement over \emph{Local}. 
This indicates that considering structural information on passages is important for the dataset.

\vspace{0.5em}
\textbf{Analysis}~~
To deal with complex questions that require evidence from multiple passages to answer, previous work \citep{wang2018evidence,lin2018denoising,wang2018joint} 
collect evidence from occurrences of an entity in different passages.
The above methods correspond to a special case of our method, i.e. MHQA with only the \emph{same}-typed edges.
From Table \ref{tab:compqa}, our method gives 1 point increase over \emph{MHQA-GRN w/ only same}, and it gives more increase in WikiHop (comparing \emph{all types} with \emph{only same} in Table \ref{tab:dev_edges}).
Both results indicate that our method could capture more useful information for multi-hop QA tasks, compared to the methods developed for previous multi-passage QA tasks.
This is likely because our method integrates not only evidences for an entity but also these for other related entities.

The leaderboard reports \emph{SplitQA} with additional sub-question annotations and gold answers for sub-questions.
These pairs of sub-questions and answers are used as additional data for training \emph{SplitQA}. 
The above approach relies on annotations of ground-truth answers for sub-questions and semantic parses, thus is not practically useful in general. 
However, the results have additional value since it can be viewed as an upper bound of \emph{SplitQA}. 
Note that the gap between this upper bound and our \emph{MHQA-GRN} is small, which further proves that larger improvement can be achieved by introducing structural information on the passage side.

\section{Related Work}

\textbf{Question answering with multi-hop reasoning}~~
%Multi-hop reasoning is an important ability for dealing with difficult cases in question answering~\citep{rajpurkar-EtAl:2016:EMNLP2016,boratko2018systematic}\mo{also adding citation to ARC, use as motivation in the introduction}. 
Most existing work on multi-hop QA focuses on hopping over knowledge bases or tables \citep{jain2016question,neelakantan2016neural,yin2016neural}, thus the problem is reduced to deduction on a readily-defined structure with known relations.
On the other hand, we study multi-hop QA on textual data and we introduce an effective approach on creating graph structures over the textual input for solving our problems.
% Here ... Different from multi-hop reading comprehension, a knowledge base is provided as the gold-standard structure for reasoning for previous multi-hop question answering.
Previous work \citep{hill2015goldilocks,shen2017reasonet} studying multi-hop QA on text does not create structures.
In addition, they only evaluate on a simple task \citep{weston2015towards} with a very limited vocabulary and passage length.
Our work is fundamentally different from theirs by modeling structures over the input, and we evaluate our models on more challenging tasks.
%However, it is still not clear how much the multi-hop reasoning abilities, such as deduction of relations and links between entities, can be reflected by the operators over single embedding vectors, neither theoretically nor empirically. For example, none of these models have been successfully verified on the recent multi-hop QA datasets.

Recent work starts to exploit ways for creating structures from inputs.
\citet{N18-1059} build a two-level computation tree over each question where the first-level nodes are sub-questions and the second-level node is a composition operation.
The answers for the sub-questions are first generated, and then combined with the composition operation.
They predefine two composition operations, so it is not general enough for other QA problems.
On the other hand, \citet{N18-2007} create DAGs over passages with coreference. 
The DAGs are then encoded with a DAG network.
Our work follows the second direction by creating graphs on the passage side.
However, we consider more types of relations than coreference, making a thorough study on relation types.
Besides, we also investigate recent graph networks on this problem.

%We observe another related Arxiv paper \citep{gcnwiki} at submission time, which is similar with ours by encoding their-created graphs with GCN.
%In addition, the edges they adopt are basically the \emph{same} and \emph{coref} edges in our approach.
%However, they only try ELMo \cite{peters2018deep} as the basis for calculating the initial context vectors for entity mentions, while all existing work use more standard models such as LSTM and GRU.
%On the other hand, our method is based on LSTM, and we have a more detailed discussion on the effectiveness of edges.

\vspace{0.2em}
\textbf{Question answering over multiple passages}
Recent efforts in open-domain QA start to generate answers from multiple passages instead of from a single passage.
%Our work is also in line with the above research direction in the sense that both aim to find answers by looking for and aggregating evidence from multiple passages.
However, most existing work on multi-passage QA selects the most relevant passage for answering the given question, thus reducing the problem to single-passage reading comprehension \citep{chen-EtAl:2017:Long4,dunn2017searchqa,dhingra2017quasar,wang2018r3,clark2018simple}.
Our method is fundamentally different by truly leveraging multiple passages.

A few multi-passage QA approaches merge evidence from multiple passages before selecting an answer \citep{wang2018evidence,lin2018denoising,wang2018joint}.
%They are similar with us in the sense that they truly utilize multiple passages and perform evidence integration.
Similar to our work, they combine evidences from multiple passages, thus they fully utilize the input passages.
%Similar to our work, the above work also deals with complex questions, which require evidence from multiple passages to answer.
The key difference is that their approaches focus on how the contexts of a single answer candidate from different passages could cover different aspects of a complex question, while our approach studies how to properly integrate the related evidence of an answer candidate, some of which come from the contexts of different entity mentions.
Specially, it increases the difficulty, since those contexts do not co-occur with the candidate answer nor the question.
%When a piece of evidence does not co-occur with the answer candidate, it is usually difficult for these methods to integrate the evidence. 
This is also demonstrated by our empirical comparison, where our approach shows much better performance than combining only the evidence of the same entity mentions.

\section{Conclusion}

We have introduced a new approach for tackling multi-hop reading comprehension (MHRC) with an evidence integration process.
Given a question and a list of passages, we first use three types of edges to connect related evidence, and then adopt recent graph neural networks to encode resulted graphs for performing evidence integration.
Results show that the three types of edges are useful on combining global evidence and that the graph neural networks are effective on encoding complex graphs resulted by the first step.
Our approach shows the highest performance among all published results on two standard MHRC datasets.

\bibliographystyle{aaai}
\bibliography{aaai}

\end{document}